# 2nd Place Solution to Facebook AI Image Similarity Challenge : Matching Track

SeungKee Jeon(Samsung Electronics)

123use321@gmail.com

## Abstract

*This paper presents the 2nd place solution to the Facebook AI Image Similarity Challenge : Matching Track on DrivenData. The solution is based on self-supervised learning, and Vision Transformer(ViT). The main breaktrough comes from concatenating query and reference image to form as one image and asking ViT to directly predict from the image if query image used reference image. The solution scored 0.8291 Micro-average Precision on the private leaderboard.*

## 1.Introduction

Facebook AI Image Similarity Challenge[1] aims to detect if reference(copyrighted) image is used in the given query image. The challenge consists of two tracks, descriptor track and matching track. In descriptor track, you're asked to represent both query and reference images in fixed(max 256) dimension embedding vector. In matching track, you're asked to directly submit corresponding reference ids and prediction scores per query. Evaluation method for both track is micro-average precision.

Micro-aveage precision is computed as follows:

$$AP = \sum_n (R_n - R_{n-1}) P_n$$

where $P_n$ and $R_n$ are the precision and recall, respectively, when thresholding at the nth image pair of the sequence sorted in order of increasing recall. This metric summarizes model performance over a range of operating thresholds from the highest predicted score to the lowest predicted score. For this challenge, the integral of the area under the curve is estimated with the finite sum over every threshold in the prediction set, without interpolation.

For this competition Facebook has compiled a new dataset, the Dataset for ISC21 (DISC21), composed of 1 million reference images and an accompanying set of 50,000 query images. Also, 1 million training images are supplied to be used for training.

## 2.Training

The training methods for both descriptor track and matching track is described in this section.

For descriptor track, I use self-supervised learning to make the model to output similar embedding features for a reference image and a query image that uses it. Vision Transformer(ViT)[2] is used as backbone network, and linear layer is followed to get 256-dimension embdding features. Training pipeline follows common self-supervised learning pipeline. Using train dataset images, augmented train images are used as query images and original images are used as reference images. For each batch, images and their augemented images are passed through the model to get their feature embeddings, and SimCLR[3] loss is used to minimize the embedding distance between correct query and reference image pair and maximize the embedding distance between wrong query and reference image pair. Detailed hyperparameters are as follows : vit_large_patch16_384, batch size 56, input image size 384x384, Adam optimizer, constant learnging rate 1e-5, SimCLR Loss.

For matching track, I concatenate query and reference image to form as one image and ask ViT to learn to predict from the image if query image used the reference image. Linear layer with output size 1 is appended after ViT to make prediction. Using train dataset images, augment train images are used as query images and original train images are used as reference images. Label is 1 if query image used reference image, and 0 if not. The model structure is depicted in Figure 1. Arguably, ViT's global attention makes it possible to learn relationship between concatenated query and reference image. Detailed hyperparameters are as follows: vit_large_patch16_224, batch size 576, concatenated image size 224x224, Adam optimizer, constant learning rate 1e-5, Binary Cross Entropy Loss.

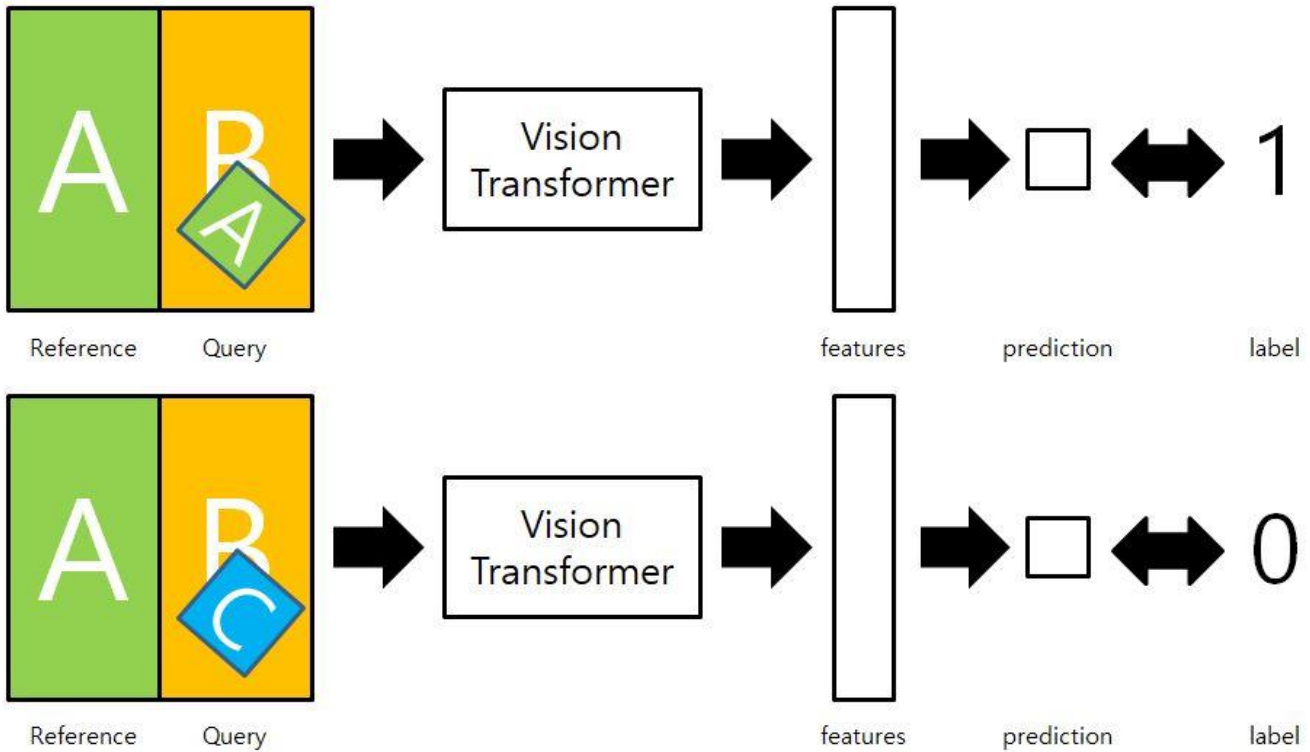

Figure 1. Matching Track model training pipeline. Reference and query image is concatenated to form as one image and ask ViT network to predict if query image used the reference image. Label is 1 if train image A is used as reference image and its augmented image is used as query image. Label is 0 if train image A is used as reference image and augmented version of other train image C is used as query image.

## 3. Inference

The inference method for matching track is described in this section.

The objective at first is to get candidate reference images for each query image. By computing cosine distance using embedding vectors, you can get closest reference images per query image. Trained model from descriptor track is used to output 256-dimesion embedding vectors for input image. Not only full image but also partial image is used to get embeddings, because sometimes reference image is used in the partial area of query image. I use three methods to get closest reference images per query image: (i) compare between full query image and full reference image (ii) compare between partial query image and full reference image (iii) compare between full query image and partial reference image. I aggregate all reference image candidates per query image from three methods to make final reference image candidates list per each query image. Finally, matching model is predicting scores for each query, reference candidate pair and reference image with highest score per query image is used for final submission.

## 4.Summary

Concatenating query and reference image to form as one image and asking ViT to directly predict from the image if query image used reference image is a simple idea, but also powerful. ViT's global attention is strong enough to learn relation between patches, even when patches include two different images, query and reference.